\pdfoutput=1

\documentclass[11pt]{article}

\usepackage{EMNLP2022}

\usepackage{times}
\usepackage{latexsym}

\usepackage[T1]{fontenc}

\usepackage[utf8]{inputenc}

\usepackage{soul}
\usepackage{url}
\usepackage{ulem}
\usepackage{booktabs}
\usepackage{amsmath}
\usepackage{amsthm} 
\usepackage{amsfonts}  
\usepackage{amssymb}
\usepackage{multicol}
\usepackage{multirow}
\usepackage{graphicx}
\usepackage{booktabs}
\usepackage{colortbl}
\usepackage{stfloats}
\usepackage{comment}
\usepackage{makecell}
\usepackage[noend]{algpseudocode}
\usepackage{algorithmicx,algorithm}
\usepackage{pifont}
\usepackage{color}
\usepackage{microtype}
\usepackage{cleveref}
\newcommand{\tabincell}[2]{\begin{tabular}{@{}#1@{}}#2\end{tabular}}
\title{Co-guiding Net: Achieving Mutual Guidances between Multiple Intent Detection and Slot Filling via Heterogeneous Semantics-Label Graphs}

\author{Bowen Xing$^{1,2}$ \and Ivor W. Tsang$^{2,1}$ \\ 
$^1$Australian Artificial Intelligence Institute, University of Technology Sydney, Australia \\
$^2$Centre for Frontier Artificial Intelligence Research, A*STAR, Singapore\\ 
\texttt{\normalsize{bwxing714@gmail.com, ivor.tsang@gmail.com}}}

\begin{document}
\maketitle
\begin{abstract}
Recent graph-based models for joint multiple intent detection and slot filling have obtained promising results through modeling the guidance from the prediction of intents to the decoding of slot filling.
However, existing methods (1) only model the \textit{unidirectional guidance} from intent to slot; (2) adopt \textit{homogeneous graphs} to model the interactions between the slot semantics nodes and intent label nodes, which limit the performance.
In this paper, we propose a novel model termed Co-guiding Net, which implements a two-stage framework achieving the \textit{mutual guidances} between the two tasks.
In the first stage, the initial estimated labels of both tasks are produced, and then they are leveraged in the second stage to model the mutual guidances.
Specifically, we propose two \textit{heterogeneous graph attention networks} working on the proposed two \textit{heterogeneous semantics-label graphs}, which effectively represent the relations among the semantics nodes and label nodes.
Experiment results show that our model outperforms existing models by a large margin, obtaining a relative improvement of 19.3\% over the previous best model on MixATIS dataset in overall accuracy.
\end{abstract} 
\section{Introduction}\label{sec:introduction}
Spoken language understanding (SLU) \cite{slu} is a fundamental task in dialog systems.
Its objective is to capture the comprehensive semantics of user utterances, and it typically includes two subtasks: intent detection and slot filling \cite{idsf}.
Intent detection aims to predict the intention of the user utterance and slot filling aims to extract additional information or constraints expressed in the utterance.

Recently, researchers discovered that these two tasks are closely tied, and a bunch of models \cite{slot-gated,selfgate,cmnet,sfid,qin2019} are proposed to combine the single-intent detection and slot filling in multi-task frameworks to leverage their correlations.

However, in real-world scenarios, a user usually expresses multiple intents in a single utterance.
To this end, \cite{kim2017} begin to tackle the multi-intent detection task and \cite{2019-joint} make the first attempt to jointly model the multiple intent detection and slot filling in a multi-task framework.
\cite{agif} propose an AGIF model to adaptively integrate the fine-grained multi-intent prediction information into the autoregressive decoding process of slot filling via graph attention network (GAT) \cite{gat}.
And \cite{glgin} further propose a non-autoregressive GAT-based model which enhances the interactions between the predicted multiple intents and the slot hidden states, obtaining state-of-the-art results and significant speedup.

\begin{figure}[t]
 \centering
 \includegraphics[width = 0.48\textwidth]{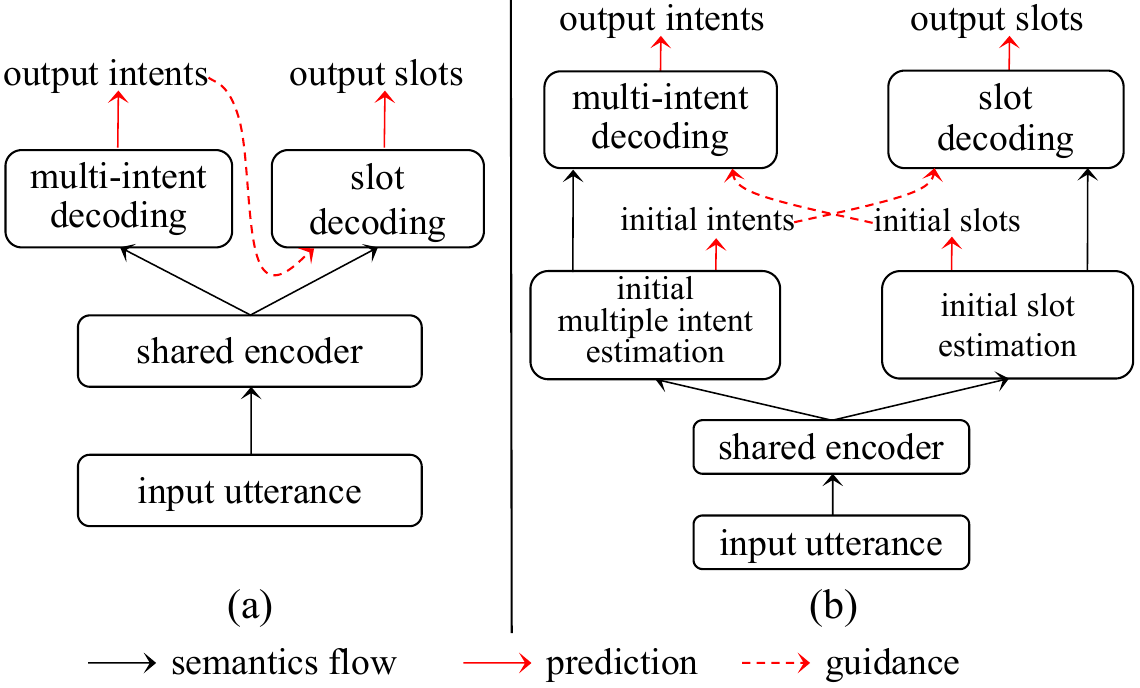}
 \caption{(a) Previous framework which only models the unidirectional guidance from multi-intent predictions to slot filling. (b) Our framework which models the mutual guidances between the two tasks.}
 \label{fig: framework}
\end{figure}

\begin{figure}[t]
 \centering
 \includegraphics[width = 0.45\textwidth]{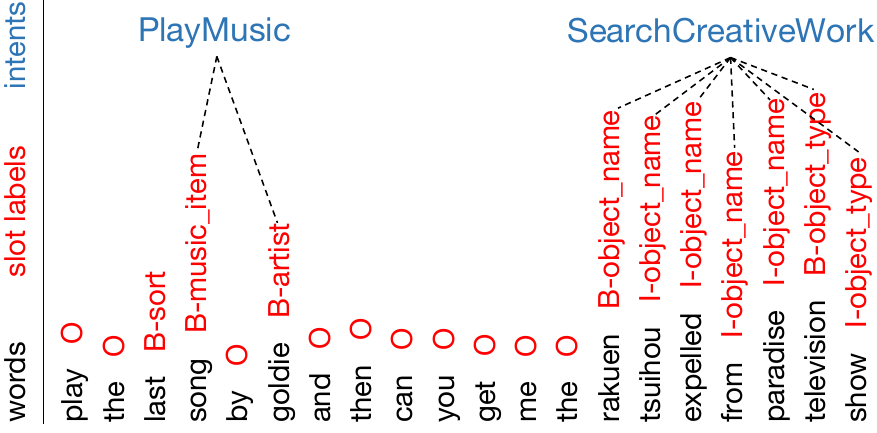}
 \caption{Illustration of the bidirectional interrelations between intent (\textcolor{blue}{blue}) and slot (\textcolor{red}{red}) labels. The sample is retrieved from MixSNIPS dataset.}
 \label{fig: example}
\end{figure}

Despite the promising progress that existing multi-intent SLU joint models have achieved, we discover that they suffer from two main issues:

(1) \textbf{Ignoring the guidance from slot to intent}. Since previous researchers realized that ``slot labels could depend on the intent'' \cite{2019-joint}, existing models leverage the information of the predicted intents to guide slot filling, as shown in Fig. \ref{fig: framework}(a).
However, they ignore that slot labels can also guide the multi-intent detection task.
Based on our observations, multi-intent detection and slot filling are bidirectionally interrelated and can mutually guide each other.
For example, in Fig \ref{fig: example}, not only the intents can indicate the slots, but also the slots
can infer the intents.
However, in previous works, 
the only guidance that the multiple intent detection task can get from the joint model is sharing the basic semantics with the slot filling task.
As a result, the lack of guidance from slot to intent limits multiple intent detection, and so the joint task.

(2) \textbf{Node and edge ambiguity in the semantics-label graph.} 
\cite{agif,glgin} apply GATs over the constructed graphs to model the interactions among the slot semantics nodes and intent label nodes.
However, their graphs are homogeneous, in which all nodes and edges are treated as the same type.
For a slot semantics node, the information from intent label nodes and other slot semantics nodes play different roles, while the homogeneous graph cannot discriminate their specific contributions, causing ambiguity.
Therefore, the heterogeneous graphs should be designed to represent the relations among the semantic nodes and label nodes to facilitate better interactions.

In this paper, we propose a novel model termed Co-guiding Net to tackle the above two issues.
For the first issue, Co-guiding Net implements a two-stage framework as shown in Fig. \ref{fig: framework} (b).
The first stage produces the initial estimated labels for the two tasks and the second stage leverages the estimated labels as prior label information to allow the two tasks mutually guide each other.
For the second issue, we propose two heterogeneous semantics-label graphs (HSLGs): (1) a slot-to-intent semantics-label graph (S2I-SLG) that effectively represents the relations among the intent semantics nodes and slot label nodes; (2) an intent-to-slot semantics-label graph (I2S-SLG) that effectively represents the relations among the slot semantics nodes and intent label nodes.
Moreover, two heterogeneous graph attention networks (HGATs) are proposed to work on the two proposed graphs for modeling the guidances from slot to intent and intent to slot, respectively.
Experiment results show that our Co-guiding Net significantly outperforms previous models, and model analysis further verifies the advantages of our model.

The contributions of our work are three-fold:
(1) We propose Co-guiding Net\footnote{https://github.com/XingBowen714/Co-guiding}, which implements a two-stage framework allowing multiple intent detection and slot filling mutually guide each other.
We make the first attempt to achieve the mutual guidances between the two tasks.
(2) We propose two heterogeneous semantics-label graphs as appropriate platforms for interactions between semantics nodes and label nodes.
And we propose two heterogeneous graph attention networks to model the mutual guidances between the two tasks.
(3) Experiment results demonstrate that our model achieves new state-of-the-art performance.
\section{Co-guiding}
\paragraph{Problem Definition}
Given a input utterance denoted as $U=\{u_i\}^n_1$,
 multiple intent detection can be formulated as a multi-label classification task that outputs multiple intent labels corresponding to the input utterance.
And slot filling is a sequence labeling task that maps each $u_i$ into a slot label.

Next, before diving into the details of Co-guiding Net's architecture, we first introduce the construction of the two heterogeneous graphs.

\subsection{Graph Construction}
\subsubsection{Slot-to-Intent Semantics-Label Graph}
\begin{figure}[t]
 \centering
 \includegraphics[width = 0.4\textwidth]{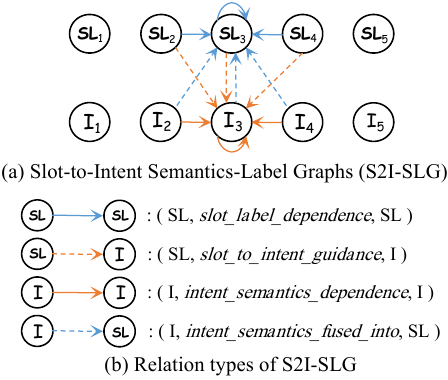}
 \caption{The illustration of S2I-SLG and its relation types. w.l.o.g, only the edges directed into SL$_3$ and I$_3$ are shown, and the local window size is 1.}
 \label{fig: S2I-SLG}
\end{figure}

To provide an appropriate platform for modeling the guidance from the estimated slot labels to multiple intent detection, we design a slot-to-intent semantics-label graph (S2I-SLG), which represents the relations among the semantics of multiple intent detection and the estimated slot labels.
S2I-SLG is a heterogeneous graph and an example is shown in Fig. \ref{fig: S2I-SLG} (a).
It contains two types of nodes: intent semantics nodes (e.g., I$_1$, ..., I$_5$) and \textbf{s}lot \textbf{l}abel (SL) nodes (e.g., SL$_1$, ..., SL$_5$).
And there are four types of edges in S2I-SLG, as shown in Fig. \ref{fig: S2I-SLG} (b).
Each edge type corresponds to an individual kind of information aggregation on the graph.

Mathematically, the S2I-SLG can be denoted as $\mathcal{G}_{s2i}=\left(\mathcal{V}_{s2i}, \mathcal{E}_{s2i}, \mathcal{A}_{s2i}, \mathcal{R}_{s2i}\right)$, in which $\mathcal{V}_{s2i}$ is the set of all nodes, $\mathcal{E}_{s2i}$ is the set of all edges, $\mathcal{A}_{s2i}$ is the set of two node types and $\mathcal{R}_{s2i}$ is the set of four edge types.
Each node $v_{s2i}$ and each edge $e_{s2i}$ are associated with their type mapping functions $\tau(v_{s2i}): \mathcal{V}_{s2i} \rightarrow \mathcal{A}_{s2i}$ and $\phi(e_{s2i}): \mathcal{E}_{s2i} \rightarrow \mathcal{R}_{s2i}$.
For instance, in Fig. \ref{fig: S2I-SLG}, the SL$_2$ node belongs to $\mathcal{V}_{s2i}$, while its node type SL belongs to $\mathcal{A}_{s2i}$; the edge from SL$_2$ to I$_3$ belongs to $\mathcal{E}_{s2i}$, while its edge type \textit{slot\_to\_intent\_guidance} belongs to $\mathcal{R}_{s2i}$.
Besides, edges in S2I-SLG are based on local connections.
For example, node I$_i$ is connected to $\{\text{I}_{i-w}, ..., \text{I}_{i+w}\}$ and $\{\text{SL}_{i-w}, ..., \text{SL}_{i+w}\}$, where $w$ is a hyper-parameter of the local window size.

\subsubsection{Intent-to-Slot Semantics-Label Graph}

\begin{figure}[t]
 \centering
 \includegraphics[width = 0.41\textwidth]{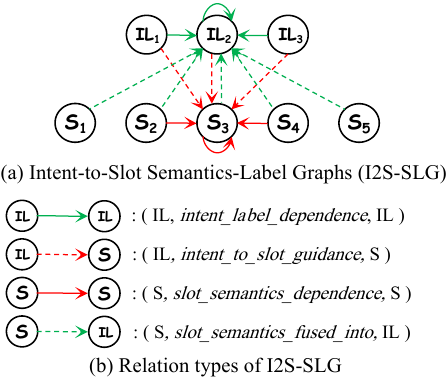}
 \caption{The illustration of I2G-SLG and its relation types. w.l.o.g, only the edges directed into IL$_3$ and S$_3$ are shown, and the local window size is 1.}
 \label{fig: I2S-SLG}
\end{figure}

To present a platform for accommodating the guidance from the estimated intent labels to slot filling, we design an intent-to-slot semantics-label graph (I2S-SLG) that represents the relations among the slot semantics nodes and the intent label nodes.
I2S-SLG is also a heterogeneous graph and an example is shown in Fig. \ref{fig: I2S-SLG} (a).
It contains two types of nodes: slot semantics nodes (e.g., S$_1$, ..., S$_5$) and \textbf{i}ntent \textbf{l}abel (IL) nodes (e.g., IL$_1$, ..., IL$_5$).
And Fig. \ref{fig: I2S-SLG} (b) shows the four edge types.
Each edge type corresponds to an individual kind of information aggregation on the graph.

Mathematically, the I2S-SLG can be denoted as $\mathcal{G}_{i2s}=\left(\mathcal{V}_{i2s}, \mathcal{E}_{i2s}, \mathcal{A}_{i2s}, \mathcal{R}_{i2s}\right)$.
Each node $v_{i2s}$ and each edge $e_{i2s}$ are associated with their type mapping functions $\tau(v_{i2s})$
and $\phi(e_{i2s})$. 
The connections in I2S-SLG are a little different from S2I-SLG.
Since intents are sentence-level, each IL node is globally connected with all nodes.
For S$_i$ node, it is connected to $\{\text{S}_{i-w}, ..., \text{S}_{i+w}\}$ and $\{\text{IL}_1, ..., \text{IL}_m\}$, where $w$ is the local window size and $m$ is the number of estimated intents.

\subsection{Model Architecture}

\begin{figure}[t]
 \centering
 \includegraphics[width = 0.48\textwidth]{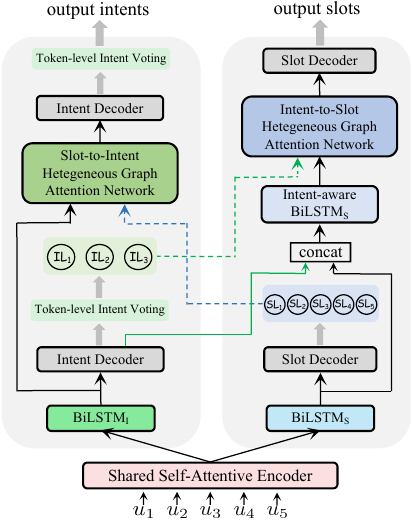}
 \caption{The architecture of Co-guiding Net. Each HGAT is triggered by its own task's semantics and the counterpart's predicted labels.
 The green and blue dashed arrow lines denote the projected label representations from the predicted intents and slots, respectively. The green solid arrow line denotes the intent distribution generated by the Intent Decoder at the first stage.}
 \label{fig: model}
\end{figure}

In this section, we introduce the details of our Co-guiding Net, whose architecture is shown in Fig.\ref{fig: model}.
\subsubsection{Shared Self-Attentive Encoder}
Following \cite{agif,glgin}, we adopt a shared self-attentive encoder to produce the initial hidden states containing the basic semantics.
It includes a BiLSTM and a self-attention module.
BiLSTM captures the temporal dependencies: 
 \begin{equation}
 \small 
h_i = \operatorname{BiLSTM} \big(x_i, h_{i-1}, h_{i+1}\big)
\end{equation}
where $x_i$ is the word vector of $u_i$.
Now we obtain the context-sensitive hidden states $\boldsymbol{\hat{H}}=\{\hat{h_i}\}_1^n$.

Self-attention captures the global dependencies: 
\begin{equation}
  \small 
\boldsymbol{H'}=\operatorname{softmax}\left(\frac{\boldsymbol{Q} \boldsymbol{K}^{\top}}{\sqrt{d_{k}}}\right) \boldsymbol{V}
\end{equation}
where $\boldsymbol{H'}$ is the global contextual hidden states output by self-attention; $\boldsymbol{Q}, \boldsymbol{K}$ and $\boldsymbol{V}$ are matrices obtained by applying different linear projections on the input utterance word vector matrix. 

Then we concatenate the output of BiLSTM and self-attention to form the output of the shared self-attentive encoder:
$\boldsymbol{H}=\boldsymbol{\hat{H}} \| \boldsymbol{H'}$, where $\boldsymbol{H}=\{h_i\}_1^n$ and $\|$ denotes concatenation operation.

\subsubsection{Initial Estimation}
\paragraph{Multiple Intent Detection}
To obtain the task-specific features for multiple intent detection, we apply a BiLSTM layer over $\boldsymbol{H}$:
 \begin{equation}
 \small 
h^{[I,0]}_i = \operatorname{BiLSTM}_{\text{I}} \left(h_i, h^{[I,0]}_{i-1}, h^{[I,0]}_{i+1}\right)
\end{equation}

Following \cite{agif,glgin}, we conduct token-level multi-intent detection.
Each $h^{[I,0]}_i$ is fed into the intent decoder. 
Specifically, the intent label distributions of the $i$-th word are obtained by:
 \begin{equation}
 \small 
y^{[I,0]}_{i}=\operatorname{sigmoid}\left(\boldsymbol{W}^1_{I}\left(\sigma(\boldsymbol{W}^2_{I} \boldsymbol{h}_{i}^{[I,0]}\!+\!\boldsymbol{b}^2_{I})\right)\!+\!\boldsymbol{b}^1_{I}\right)
\end{equation}
where $\sigma$ denotes the non-linear activation function; $W_*$ and $b_*$ are model parameters.

Then the estimated sentence-level intent labels $\{\text{IL}_1, ..., \text{IL}_m\}$ are obtained by the token-level intent voting \cite{glgin}.
\paragraph{Slot Filling}
\cite{glgin} propose a non-autoregressive paradigm for slot filling decoding, which achieves significant speedup.
In this paper, we also conduct parallel slot filling decoding.

We first apply a BiLSTM over $\boldsymbol{H}$ to obtain the task-specific features for slot filling:
\begin{equation}
 \small 
h^{[S,0]}_i = \operatorname{BiLSTM}_{\text{S}} (h_i, h^{[S,0]}_{i-1}, h^{[S,0]}_{i+1})
\end{equation}
Then use a softmax classifier to generate the slot label distribution for each word:
\begin{equation}
 \small 
y_{i}^{[S,0]}=\operatorname{softmax} \left(\boldsymbol{W}^1_{S}\left(\sigma (\boldsymbol{W}^2_{S} \boldsymbol{h}_{i}^{[S,0]}\!+\!\boldsymbol{b}^2_{S})\right)\!+\!\boldsymbol{b}^1_{S}\right)
\end{equation}
And the estimated slot label for each word is obtained by $\text{SL}_i = \operatorname{arg\ max}(y_{i}^{[S,0]})$.

\subsubsection{Heterogeneous Graph Attention Network}
State-of-the-art models \cite{agif,glgin} use a homogeneous graph to connect the semantic nodes of slot filling and the intent label nodes.
And GAT \cite{gat} is adopted to achieve information aggregation.
In Sec. \ref{sec:introduction}, we propose that this manner cannot effectively learn the interactions between one task's semantics and the estimated labels of the other task.
To tackle this issue, we propose two heterogeneous graphs (S2I-SLG and I2S-SLG) to effectively represent the relations among the semantic nodes and label nodes.
To model the interactions between semantics and labels on the proposed graphs, we propose a Heterogeneous Graph Attention Network (HGAT).
When aggregating the information into a node, HGAT can discriminate the specific information from different types of nodes along different relations.
And two HGATs (S2I-HGAT and I2S-HGAT) are applied on S2I-SLG and I2S-SLG, respectively.
Specifically, 
 S2I-HGAT can be formulated as follows: 
\begin{equation}
 \small 
 \begin{split}
&h^{l+1}_i \!= \! \mathop{\|}\limits_{k=1}^K\sigma\!\left(\sum_{j\in \mathcal{N}_{s2i}^{i}} \!\!\! W_{s2i}^{[r,k,1]} \alpha_{ij}^{[r,k]} h^{l}_j \!\right)\!,  r = \phi\!\left(e^{[j,i]}_{s2i}\right)\\
&\alpha_{ij}^{[r,k]}\! = \!\frac{\operatorname{exp} \!\left( \! \left( \!W_{s2i}^{[r,k,2]} h^{l}_i \! \right) \! \left( \!W_{s2i}^{[r,k,3]} h^{l}_j \!\right)^\mathsf{T} \!\!/\! \sqrt d  \right)}{\sum\limits_{u\in\mathcal{N}_{s2i}^{r,i}} \! \operatorname{exp} \! \left( \! \left( \!W_{s2i}^{[r,k,2]} h^{l}_i \!\right) \! \left( \!W_{s2i}^{[r,k,3]} h^{l}_u \!\right)^\mathsf{T} \!\!\!/ \!\sqrt d \right)}
\end{split}\label{eq: hgat}
\end{equation}
where $K$ denotes the total head number; $\mathcal{N}_{s2i}^{i}$ denotes the set of incoming neighbors of node $i$ on S2I-SLG; $W_{s2i}^{[r,k,*]}$ are weight matrices of edge type $r$ on the $k$-th head; $e^{[j,i]}_{s2i}$ denotes the edge from node $j$ to node $i$ on S2I-SLG; $\mathcal{N}_{s2i}^{r,i}$ denotes the nodes connected to node $i$ with $r$-type edges on S2I-SLG; $d$ is the dimension of node hidden state.

I2S-HGAT can be derived like Eq. \ref{eq: hgat}.
\subsubsection{Intent Decoding with Slot Guidance}
In the first stage, we obtain the initial intent features $H^{[I,0]}=\{h^{I,0}_i\}_i^n$ and the initial estimated slot labels sequence $\{\text{SL}_1, ..., \text{SL}_n\}$.
Now we project the slot labels into vector form using the slot label embedding matrix, obtaining $E_{sl}=\{e^1_{sl}, ..., e^n_{sl}\}$.

Then we feed $H^{[I,0]}$ and $E_{sl}$ into S2I-HGAT to model their interactions, allowing the estimated slot label information to guide the intent decoding:
\begin{equation}
 \small 
 H^{[I,L]}=\operatorname{S2I-HGAT}\left([H^{[I,0]}, E_{sl}],\mathcal{G}_{s2i}, \theta_I\right)
\end{equation}
where $[H^{[I,0]}, E_{sl}]$ denotes the input node representation; $\theta_I$ denotes S2I-HGAT's parameters.
$L$ denotes the total layer number.

Finally, $H^{[I,L]}$ is fed to intent decoder, producing the intent label distributions for the utterance words: $Y^{[I,1]}=\{y^{[I,1]}_i, ..., y_n^{[I,1]}\}$.
And the final output sentence-level intents are obtained via applying token-level intent voting over $Y^{[I,1]}$.
\subsubsection{Slot Decoding with Intent Guidance}
\paragraph{Intent-aware BiLSTM} Since the B-I-O tags of slot labels have temporal dependencies, we use an intent-aware BiLSTM to model the temporal dependencies among slot hidden states with the guidance of estimated intents:
\begin{equation}
 \small 
 \tilde{h}_i^{[S,0]}=\operatorname{BiLSTM}(y_i^{[I,0]}\|h_i^{[S,0]}, \tilde{h}_{i-1}^{[S,0]}, \tilde{h}_{i+1}^{[S,0]})
\end{equation}

\paragraph{I2S-HGAT} We first project the estimated intent labels $\{\text{IL}_j\}_1^m$ into vectors using the intent label embedding matrix, obtaining $E_{il}=\{e^1_{il}, ..., e^m_{il}\}$.
Then we feed $\tilde{H}^{S}$ and $E_{il}$ into I2S-HGAT to model their interactions, allowing the estimated intent label information to guide the slot decoding:
\begin{equation}
 \small 
 H^{[S,L]}=\operatorname{I2S-HGAT}\left([\tilde{H}^{S}, E_{il}],\mathcal{G}_{i2s}, \theta_S\right)
\end{equation}
where $[\tilde{H}^{[S]}, E_{il}]$ denotes the input node representation; $\theta_S$ denotes I2S-HGAT's parameters.

Finally, $H^{[S,L]}$ is fed to slot decoder, producing the slot label distributions for each word: $Y^{[S,1]}=\{y^{[S,1]}_i, ..., y_n^{[S,1]}\}$.
And the final output slot labels are obtained by applying $\operatorname{arg\ max}$ over $Y^{[S,1]}$.
\subsubsection{Training Objective}
\paragraph{Loss Function}
The loss function for multiple intent detection is:
\begin{equation}
 \small 
 \begin{split}
\mathrm{CE}(\hat{y}, y)&=\hat{y} \log (y)+(1-\hat{y}) \log (1-y)\\
\mathcal{L}_{I} &=\sum_{t=0}^1\sum_{i=1}^{n} \sum_{j=1}^{N_{I}} \mathrm{CE}\left(\hat{y}_{i}^{I}[j], y_{i}^{[I,t]}[j]\right)
\end{split}
\end{equation}
And the loss function for slot filling is:
\begin{equation}
 \small 
\mathcal{L}_{S} = \sum_{t=0}^1\sum_{i=1}^{n} \sum_{j=1}^{N_{S}} \hat{y}_{i}^{S}[j] \log \left(y_{i}^{[S,t]}[j]\right)
\end{equation}
where ${N}_I$ and ${N}_S$ denote the total numbers of intent labels and slot labels; $\hat{y}_{i}^{I}$ and $\hat{y}_{i}^{S}$ denote the ground-truth intent labels and slot labels. 
\paragraph{Margin Penalty}
The core of our model is to let the two tasks mutually guide each other.
Intuitively, the predictions in the second stage should be better than those in the first stage.
To force our model obey this rule, we design a margin penalty ($\mathcal{L}^{mp}$) for each task, whose aim is to improve the probabilities of the correct labels. Specifically, the formulations of $\mathcal{L}_I^{mp}$ and $\mathcal{L}_S^{mp}$ are:
\begin{equation}
\small
\setlength{\abovedisplayskip}{7pt} 
 \begin{split}
&\mathcal{L}_I^{mp} \!=\! \sum_{i=1}^n \sum_{j=1}^{{N}_I} \hat{y}_{i}^{I}[j]\ \operatorname{max} \left(0, y_i^{[I,0]}[j]-y_i^{[I,1]}[j]\right)\\
&\mathcal{L}_S^{mp} \!=\! \sum_{i=1}^{n} \sum_{j=1}^{N_{S}} \hat{y}_{i}^{S}[j] \operatorname{max}  \left(0, y_i^{[S,0]}[j]-y_i^{[S,1]}[j]\right)
\end{split}
\end{equation}
\paragraph{Model Training}
The training objective $\mathcal{L}$ is the weighted sum of loss functions and margin regularizations of the two tasks: 
\begin{equation}
 \small 
\mathcal{L} = \gamma\left(\mathcal{L}_I + \beta_I\mathcal{L}_I^{mp}\right) + \left(1-\gamma\right)\left(\mathcal{L}_S + \beta_S \mathcal{L}_S^{mp}\right) \label{eq: final objective}
\end{equation}
where $\gamma$ is the coefficient balancing the two tasks; $\beta_I$ and $\beta_S$ are the coefficients of the margin regularization for the two tasks.
\section{Experiments}
\subsection{Datasets and Metrics}
Following previous works, MixATIS and MixSNIPS \cite{atis, snips,agif} are taken as testbeds.
MixATIS includes 13,162 utterances for training, 756 ones for validation and 828 ones for testing.
MixSNIPS includes 39,776 utterances for training, 2,198 ones for validation and 2,199 ones for testing.

As for evaluation metrics, following previous works, we adopt accuracy (Acc) for multiple intent detection, F1 score for slot filling, and overall accuracy for the sentence-level semantic frame parsing.
Overall accuracy denotes the ratio of sentences whose intents and slots are all correctly predicted.

\subsection{Implementation Details}
Following previous works, the word and label embeddings are trained from scratch\footnote{Due to space limitation, the experiments using pre-trained language model as the encoder are presented in Appendix.}.
The dimensions of word embedding, label embedding, and hidden state are 256 on MixATIS, while on MixSNIPS they are 256, 128, and 256.
The layer number of all GNNs is 2.
Adam \cite{adam} is used to train our model with a learning rate of $1e^{-3}$ and a weight decay of $1e^{-6}$.
As for the coefficients Eq.\ref{eq: final objective}, $\gamma$ is 0.9 on MixATIS and 0.8 on MixSNIPS; on both datasets, $\beta_I$ is $1e^{-6}$ and $\beta_S$ is $1e^0$.
The model performing best on the dev set is selected then we report its results on the test set.
All experiments are conducted on RTX 6000.
Our source code will be released.

\subsection{Main Results}
\begin{table*}[t]
\centering
\fontsize{8}{10}\selectfont
\setlength{\tabcolsep}{0.8mm}{
\begin{tabular}{l|ccc|ccc}
\toprule
\multirow{2}{*}{Models} & \multicolumn{3}{c|}{MixATIS} & \multicolumn{3}{c}{MixSNIPS} \\ \cline{2-7} 
                        & Overall(Acc)  &Slot (F1)  &Intent(Acc)& Overall(Acc)& Slot(F1)&Intent(Acc)           \\ \midrule
Attention BiRNN \cite{attbirnn} &  39.1 &  86.4      &   74.6      & 59.5        &  89.4     & 95.4 \\
Slot-Gated \cite{slot-gated}    &  35.5 &  87.7      &   63.9      & 55.4        &  87.9     & 94.6 \\
Bi-Model \cite{bimodel}         &  34.4 &  83.9      &   70.3      & 63.4        &  90.7    & 95.6 \\
SF-ID \cite{sfid}               &  34.9 &  87.4      &   66.2      & 59.9        &  90.6     & 95.0 \\
Stack-Propagation \cite{qin2019}&  40.1 &  87.8      &   72.1      & 72.9        &  94.2     & 96.0 \\
Joint Multiple ID-SF \cite{2019-joint}&36.1 &84.6    &   73.4      & 62.9        &  90.6     & 95.1 \\
AGIF \cite{agif}                &  40.8 &  86.7      &   74.4      & 74.2        &  94.2     & 95.1 \\
GL-GIN \cite{glgin}        &  43.0 &  88.2      &   76.3      & 73.7        &  94.0     & 95.7 \\ \midrule
Co-guiding Net (ours)           &  \textbf{51.3}$^\dag$ &\textbf{89.8}$^\dag$ &\textbf{79.1}$^\dag$ & \textbf{77.5}$^\dag$  &  \textbf{95.1}$^\dag$ & \textbf{97.7}$^\dag$ \\
  \bottomrule
\end{tabular}}
\caption{Results comparison. $^\dag$ denotes our model significantly outperforms baselines with $p<0.01$ under t-test.} 
\label{table: main results}
\end{table*}

The performance comparison of Co-guiding Net and baselines are shown in Table \ref{table: main results}, from which we have the following observations:

(1) Co-guiding Net gains significant and consistent improvements on all tasks and datasets.
Specifically, 
on MixATIS dataset, it overpasses the previous state-of-the-art model GL-GIN by 19.3\%, 1.8\%, and 3.7\% on sentence-level semantic frame parsing, slot filling, and multiple intent detection, respectively;
on MixSNIPS dataset, it overpasses GL-GIN by 5.2\%, 1.2\% and 2.1\% on sentence-level semantic frame parsing, slot filling and multiple intent detection, respectively.
This is because our model achieves the mutual guidances between multiple intent detection and slot filling, allowing the two tasks to provide crucial clues for each other.
Besides, our designed HSLGs and HGATs can effectively model the interactions among the semantics nodes and label nodes, extracting the indicative clues from initial predictions.

(2) Co-guiding Net achieves a larger improvement on multiple intent detection than slot filling.
The reason is that except for the guidance from multiple intent detection to slot filling, our model also achieves the guidance from slot filling to multiple intent detection, while previous models all ignore this.
Besides, previous methods model the semantics-label interactions by homogeneous graph and GAT, limiting the performance.
Differently, our model uses the heterogeneous semantics-label graphs to represent different relations among the semantic nodes and the label nodes, then applies the proposed HGATs over the graphs to achieve the interactions.
Consequently, their performances (especially on multiple intent detection) are significantly inferior to our model.

(3) The improvements in overall accuracy are much sharper.
We suppose the reason is that the achieved mutual guidances make the two tasks deeply coupled and allow them to stimulate each other using their initial predictions.
For each task, its final outputs are guided by its and another task's initial predictions.
By this means, the correct predictions of the two tasks can be better aligned.
As a result,  more test samples get correct sentence-level semantic frame parsing results, and then overall accuracy is boosted.

\subsection{Model Analysis}

\begin{table*}[t]
\centering
\fontsize{9}{11}\selectfont
\setlength{\tabcolsep}{2.5mm}{
\begin{tabular}{l|ccc|ccc}
\toprule
\multirow{2}{*}{Models} & \multicolumn{3}{c|}{MixATIS} & \multicolumn{3}{c}{MixSNIPS} \\ \cline{2-7} 
                        & Overall(Acc)  &Slot (F1)  &Intent(Acc)& Overall(Acc)& Slot(F1)&Intent(Acc)           \\ \midrule
Co-guiding Net          &  \textbf{51.3} &\textbf{89.8} &\textbf{79.1} & \textbf{77.5}  &  \textbf{95.1} & \textbf{97.7} \\\midrule
w/o  S2I-guidance  &  47.7 ($\downarrow$3.6) &  88.8 ($\downarrow$1.0)      &   77.1    ($\downarrow$2.0)  & 76.6 ($\downarrow$0.9)       &  94.7 ($\downarrow$0.4)    & 96.9 ($\downarrow$0.8) \\
w/o  I2S-guidance &  47.7 ($\downarrow$3.6)&  88.7  ($\downarrow$1.1)    &   77.5  ($\downarrow$1.6)    & 76.5    ($\downarrow$1.0)    &  94.9  ($\downarrow$0.2)  & 97.5 ($\downarrow$0.2)\\
w/o relations &  46.0 ($\downarrow$5.3) &  88.3  ($\downarrow$1.5)  & 77.8 ($\downarrow$1.3)   & 76.3       ($\downarrow$1.2) &  94.7    ($\downarrow$0.5) & 97.2 ($\downarrow$0.4) \\
+ Local Slot-aware GAT & 51.1  ($\downarrow$0.2) &  89.4  ($\downarrow$0.4)  &   79.0 ($\downarrow$0.1) & 75.9   ($\downarrow$1.6)     &  94.7 ($\downarrow$0.4)    & 96.4 ($\downarrow$1.4)  \\
  \bottomrule
\end{tabular}}
\caption{Results of ablation experiments.} 
\label{table: ablation}
\end{table*}
We conduct a set of ablation experiments to verify the advantages of our work from different perspectives, and the results are shown in Table \ref{table: ablation}.

\subsubsection{Effect of Slot-to-Intent Guidance}
One of the core contributions of our work is achieving the mutual guidances between multiple intent detection and slot filling, while previous works only leverage the one-way message from intent to slot.
Therefore, compared with previous works, one of the advantages of our work is modeling the slot-to-intent guidance.
To verify this, we design a variant termed \textit{w/o S2I-guidance} and its result is shown in Table \ref{table: ablation}.
We can observe that Intent Acc drops by 2.0\% on MixATIS and 0.8\% on MixSNIPS.
Moreover, Overall Acc drops more significantly: 3.6\% on MixATIS and 0.9\% on MixSNIPS.
This proves that the guidance from slot to intent can effectively benefit multiple intent detection, and achieving the mutual guidances between the two tasks can significantly improve Overall Acc.

Besides, although both of \textit{w/o S2I-guidance} and GL-GIN only leverage the one-way message from intent to slot, \textit{w/o S2I-guidance} outperforms GL-GIN by large margins.
We attribute this to our proposed heterogeneous semantics-label graphs and heterogeneous graph attention networks, whose advantages are verified in Sec. \ref{sec: hslghgat}.

\subsubsection{Effect of Intent-to-Slot Guidance}
To verify the effectiveness of intent-to-slot guidance, we design a variant termed \textit{w/o I2S-guidance} and its result is shown in Table \ref{table: ablation}.
We can find that the intent-to-slot guidance has a significant impact on performance.
Specifically, \textit{w/o I2S-guidance} cause nearly the same extent of performance drop on Overall Acc, proving that both of the intent-to-slot guidance and slot-to-intent guidance are indispensable and achieving the mutual guidances can significantly boost the performance.

\subsubsection{Effect of HSLGs and HGATs} \label{sec: hslghgat}
In this paper, we design two HSLGs: (i.e., S2I-SLG, I2S-SLG) and two HGATs (i.e., S2I-HGAT, I2S-HGAT).
To verify their effectiveness, we design a variant termed \textit{w/o relations} by removing the relations on the two HSLGs.
In this case, S2I-SLG/I2S-SLG collapses to a homogeneous graph, and S2I-HGAT/I2S-HGAT collapses to a general GAT based on multi-head attentions.
From Table \ref{table: ablation}, we can observe that \textit{w/o relations} obtains dramatic drops on all metrics on both datasets. 
The apparent performance gap between \textit{w/o relations} and Co-guiding Net verifies that (1) our proposed HSLGs can effectively represent the different relations among the semantics nodes and label nodes, providing appropriate platforms for modeling the mutual guidances between the two tasks; (2) our proposed HGATs can sufficiently and effectively model interactions between the semantics and indicative label information via achieving the relation-specific attentive information aggregation on the HSLGs.

Besides, although \textit{w/o relations} obviously underperforms Co-guiding Net, it still significantly outperforms all baselines.
We attribute this to the fact that our model achieves the mutual guidances between the two tasks, which allows them to promote each other via cross-task correlations.

\subsubsection{Effect of I2S-HGAT for Capturing Local Slot Dependencies}
\citet{glgin} propose a Local Slot-aware GAT module to alleviate the uncoordinated slot problem (e.g., \textit{B-singer} followed by \textit{I-song}) \cite{slotrefine} caused by the non-autoregressive fashion of slot filling.
And the ablation study in \cite{glgin} proves that this module effectively improves the slot filling performance by modeling the local dependencies among slot hidden states.
In their model (GL-GIN), the local dependencies are modeled in both of the local slot-aware GAT and subsequent global intent-slot GAT.
We suppose the reason why GL-GIN needs the local Slot-aware GAT is that the global intent-slot GAT in GL-GIN cannot effectively capture the local slot dependencies.
GL-GIN's global slot-intent graph is homogeneous, and the GAT working on it treats the slot semantics nods and the intent label nodes equally without discrimination.
Therefore, each slot hidden state receives indiscriminate information from both of its local slot hidden states and all intent labels, making it confusing to capture the local slot dependencies.
In contrast, we believe our I2S-HLG and I2S-HGAT can effectively capture the slot local dependencies along the specific \textit{slot\_semantics\_dependencies} relation, which is modeled together with other relations.
Therefore, our Co-guiding Net does not include another module to capture the slot local dependencies.

To verify this, we design a variant termed \textit{+Local Slot-aware GAT}, which is implemented by augmenting Co-guiding Net with the Local Slot-aware GAT \cite{glgin} located after the Intent-aware BiLSTM$_s$ (the same position with GL-GIN).
And its result is shown in Table \ref{table: ablation}.
We can observe that not only the Local Slot-aware GAT does not bring improvement, it even causes performance drops.
This proves that our I2S-HGAT can effectively capture the local slot dependencies.

\subsection{Case Study}
\begin{figure*}[t]
 \centering
 \includegraphics[width = 0.98\textwidth]{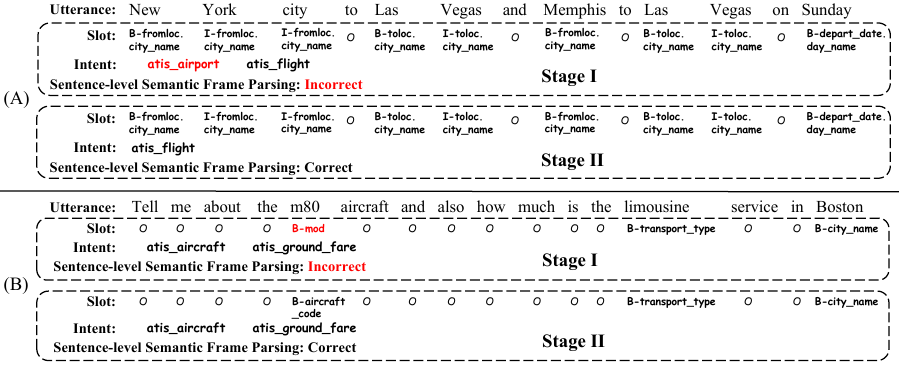}
 \caption{Case study of slot-to-intent guidance (A) and intent-to-slot guidance (B).  Red color denotes error.}
 \label{fig: case study}
\end{figure*}

To demonstrate how our model allows the two tasks to guide each other, we present two cases in Fig. \ref{fig: case study}.

\paragraph{Slot-to-Intent Guidance}
From Fig. \ref{fig: case study} (A), we can observe that in the first stage, all slots are correctly predicted, while multiple intent detection obtains a redundant intent \texttt{atis\_airport}.
In the second stage, our proposed S2I-HGAT operates on S2I-HLG.
It aggregates and analyzes the slot label information from the slot predictions of the first stage, extracting the indicative information that most slot labels are about \texttt{city\_name} while no information about \textit{airport} is mentioned.
Then this beneficial guidance information is passed into intent semantics nodes whose representations are then fed to the intent decoder for prediction.
In this way, the guidance from slot filling helps multiple intent detection predict correctly.

\paragraph{Intent-to-Slot Guidance}
In the example shown in Fig. \ref{fig: case study} (B), in the first stage, correct intents are predicted, while there is an error in the predicted slots.
In the second stage, our proposed I2S-HGAT operates on I2S-HLG.
It comprehensively analyzes the indicative information of \textit{airecraft} from both of slot semantics node \textit{aircraft} and intent label node \texttt{atis\_aircraft}.
Then this beneficial guidance information is passed into the slot semantics of \textit{m80},
whose slot is therefore correctly inferred.
\section{Related Work}\label{sec: relatedwork}
The correlations between intent detection and slot filling have been widely recognized.
To leverage them, a group of models \cite{ijcai2016joint,hakkani2016multi,slot-gated,selfgate,sfid,cmnet,qin2019,jointcap,slotrefine,qin2021icassp,ni2021recent} are proposed to tackle the joint task of intent detection and slot filling in a multi-task manner.
However, the intent detection modules in the above models can only handle the utterances expressing a single intent, which may not be practical in real-world scenarios, where there are usually multi-intent utterances.

To this end, \citet{kim2017} propose a multi-intent SLU model, and \cite{2019-joint} propose the first model to jointly model the tasks of multiple intent detection and slot filling via a slot-gate mechanism.
Furthermore, as graph neural networks have been widely utilized in various tasks \cite{cao-etal-2019-multi,rgat,shi-etal-2021-transfernet,tetci,darer,dignet}, they have been leveraged to model the correlations between intent and slot.
\citet{agif} propose an adaptive graph-interactive framework to introduce the fine-grained multiple intent information into slot filling achieved by GATs.
More recently, \citet{glgin} propose another GAT-based model, which includes a non-autoregressive slot decoder conducting parallel decoding for slot filling and achieves the state-of-the-art performance.

Our work also tackles the joint task of multiple intent detection and slot filling.
Existing methods only model the one-way guidance from multiple intent detection to slot filling.
Besides, they adopt homogeneous graphs and vanilla GATs to achieve the interactions between the predicted intents and slot semantics.
In contrast, we (1) achieve the mutual guidances between the two tasks; (2)  propose the heterogeneous semantics-label graphs to represent the dependencies among the semantics and predicted labels; (3) we propose the Heterogeneous Graph Attention Network to model the semantics-label interactions on the heterogeneous semantics-label graphs.

\section{Conclusion}\label{sec: conclusion}
In this paper, 
we propose a novel Co-guiding Net based on a two-stage framework that allows the two tasks to guide each other in the second stage using the predicted labels at the first stage.
To represent the relations among the semantics node and label nodes, we propose two heterogeneous semantics-label graphs, and two heterogeneous graph attention networks are proposed to model the mutual guidances between intents and slots.
Experiment results on benchmark datasets show that our model significantly outperforms previous models.
Future work will focus on leveraging syntactic information to enhance utterance understanding.

\section*{Limitations}
Although our Co-guiding Net achieves significant improvement over existing models, we suppose that its sentence understanding module (the self-attentive encoder) is not sufficient enough and limits the performance to some extent.
In recent years, the syntactic information extracted from the sentence's syntax/dependency tree, which is output by an off-the-shelf dependency parser, has been widely leveraged to assist the models in comprehensively understanding the sentence \cite{rgat,coreference,re-agcn,neuralsubgraph,jair}.
And in SLU, each utterance is a sentence, and we believe that leveraging the syntactic information can improve the performance by enriching the word representations.
\section*{Acknowledgements}
This work was supported by Australian Research Council  Grant DP200101328.
Bowen Xing and Ivor W. Tsang were also supported by A$^*$STAR Centre for Frontier AI Research.
\normalem
\bibliography{anthology}
\bibliographystyle{acl_natbib}

\appendix

\section{Experiments using Pre-trained Language Model}
\subsection{Settings}
To evaluate Co-guiding Net's performance based on the pre-trained language model, we use the pre-trained RoBERTa \cite{roberta} encoder to replace the original self-attentive encoder.
We adopt the pre-trained RoBERTa-base version provided by Transformers \cite{transformers}.
For each word, its first subwords' hidden state generated by RoBERTa is taken as the word representation.
AdamW \cite{weightdecay} optimizer is used for model training with the default setting, and RoBERTa is fine-tuned with model training. 
Other model components are identical to the Co-guiding Net based on LSTM, and we use the same hyper-parameters of the model rather than search for the optimal ones for RoBERTa+Co-guiding Net due to our limited computation resource.

\subsection{Results}
Table \ref{table: roberta} shows the result comparison of Co-guiding Net, RoBERTa+Co-guiding Net, and their state-of-the-art counterparts: AGIF, GL-GIN, RoBERTa+AGIF, and RoBERTa+GL-GIN.
We can find that although RoBERTa boosts the models' performance, RoBERTa+Co-guiding Net still significantly outperforms RoBERTa+AGIF and RoBERTa+GL-GIN.
This can be attributed to the fact that although the pre-trained language model (PTLM) can enhance the word representations, it cannot achieve the guidance between the two tasks or the interactions between the semantics and label information, which are exactly the advantages of our Co-guiding Net.
Therefore, collaborating with PTLM that has strong ability of language modeling, RoBERTa+Co-guiding Net gets its performance further boosted, achieving new state-of-the-art.
\begin{table}[t]
\centering
\fontsize{8}{9}\selectfont
\setlength{\tabcolsep}{0.8mm}{
\begin{tabular}{l|c|c}
\toprule
\multirow{2}{*}{Models} & \multicolumn{1}{c|}{MixATIS} & \multicolumn{1}{c}{MixSNIPS} \\ \cline{2-3} 
                        & Overall(Acc) & Overall(Acc)       \\ \midrule
 AGIF        &  40.8  & 74.2  \\                   
GL-GIN      &  43.0  & 73.7  \\ 
Co-guiding Net (ours)           &  \textbf{51.3}$^\dag$ & \textbf{77.5}$^\dag$ \\ \midrule
 RoBERTa+AGIF        &  50.0  & 80.7 \\                   
RoBERTa+GL-GIN      &  53.6  & 82.6  \\ 
RoBERTa+Co-guiding Net (ours)           &  \textbf{57.5}$^\dag$ & \textbf{85.3}$^\dag$ \\ \midrule
\end{tabular}}
\caption{Results comparison. $^\dag$ denotes our model significantly outperforms the corresponding counterparts with $p<0.01$ under t-test.} 
\label{table: roberta}
\end{table}

\label{sec:appendix}


\end{document}